# Developing A Fair Individualized Polysocial Risk Score (iPsRS) for Identifying Increased Social Risk of Hospitalizations in Patients with Type 2 Diabetes (T2D)

Yu Huang[1a], Jingchuan Guo[2a], William T Donahoo[3], Zhengkang Fan[1], Ying Lu[2], Wei-Han Chen[2], Huilin Tang[2], Lori Bilello[4], Elizabeth A Shenkman[1], Jiang Bian[1]

[1] Department of Health Outcomes and Biomedical Informatics, University of Florida, Gainesville, FL, USA

[2] Pharmaceutical Outcomes & Policy, University of Florida, Gainesville, FL, USA

[3] Division of Endocrinology, Diabetes and Metabolism, University of Florida College of Medicine

[4] Department of Medicine, University of Florida College of Medicine

[a] Contributed equally, co-first author

Corresponding authors:

Jiang Bian, PhD bianjiang@ufl.edu

Affiliation: Department of Health Outcomes & Biomedical Informatics, University of Florida

Address: 2197 Mowry Road, 122 PO Box 100177 Gainesville, FL 32610-0177

Phone Number: (352) 273-8878

**Abstract**

**Background:** Racial and ethnic minority groups and individuals facing social disadvantages, which often stem from their social determinants of health (SDoH), bear a disproportionate burden of type 2 diabetes (T2D) and its complications. It is therefore crucial to implement effective social risk management strategies at the point of care.

**Objective:** To develop EHR-based machine learning (ML) analytical pipeline to identify the unmet social needs associated with hospitalization risk in patients with T2D.

**Methods:** We identified real-world patients with T2D from the electronic health records (EHR) data from University of Florida (UF) Health Integrated Data Repository (IDR), incorporating both contextual SDoH (e.g., neighborhood deprivation) and individual-level SDoH (e.g., housing instability). Our 2012-2020 data were used for training and validation and 2021-2022 data for independent testing. We developed a ML analytic pipeline, namely individualized polysocial risk score (iPsRS), to identify high social risk associated with hospitalizations in T2D patients, along with explainable AI (XAI, e.g., SHAP and causal learning) and fairness assessment and optimization.

**Results:** The study cohort included 10,192 real-world patients with T2D, with a mean age of 59 years and 58% female. Of the cohort, 50% were non-Hispanic White, 39% were non-Hispanic Black, 6% were Hispanic, and 5% were other races/ethnicities. Our iPsRS, including both contextual and individual-level SDoH as input factors, achieved a C statistic of 0.72 in predicting 1-year hospitalization after fairness optimization across racial and ethnic groups. The iPsRS showed excellent utility for capturing individuals at high hospitalization risk because of SDoH, that is, the actual 1-year hospitalization rate in the top 5% of iPsRS was 28.1%, ~13 times as high as the bottom decile (2.2% for 1-year hospitalization rate). In a multiple logistic regression model adjusting for patients' demographic and clinical characteristics (e.g., age, sex, and comorbidities), our iPsRS explained 40.3% of the risk of 1-year hospitalization; per decile increase in iPsRS, the hospitalization risk increased by 16% (adjusted odds ratio=1.16, 95%CI 1.10-1.22).

**Conclusion:** Our ML pipeline iPsRS can fairly and accurately screen for patients who have increased social risk leading to hospitalization in real word patients with T2D.

**Introduction**

Diabetes affects 529 million people worldwide and the number is projected to more than double in the next three decades, reaching 1.3 billion by 2050.[1] Over 90% of diabetes cases are type 2 diabetes (T2D).[2] Existing research has shown that social determinants of health (SDoH)—"*the conditions in the environments where people are born, live, learn, work, play, worship, and age,*"[3,4] such as education, income, lifestyle choices, and access to healthy food, play a critical role affecting a wide range of health, functioning, and disease risk, including in the development and prognosis of T2D.[5–7] Moreover, health disparities in T2D are widely documented over the past decades.[8–10] Racial and ethnic minority groups and individuals experiencing social disadvantages—often rooted in their SDoH—bear a disproportionate burden of T2D and its complications.[11–13] As such, diabetes is a public crisis that must be managed by addressing patients' unmet social needs, especially for marginalized groups, to improve T2D outcomes and health equity.

Effective social risk management strategies are essential to the improvement of T2D outcomes and health equity. The US health care system has begun embracing patients' social needs, including screening for SDoH at the point of care. Payers have also started to consider their beneficials' unmet social needs and their SDoH. For example, the Centers for Medicare & Medicaid Services (CMS) have made various proposals to incorporating health-related social needs (HRSNs) into the Medicare and Medicaid programs[14] such as requiring SDoH (e.g., housing stability, food insecurity, and access to transportation) to be included in annual beneficiary health risk assessments. However, only 16%-24% of clinics and hospitals provide SDoH screening,[15] and the actual utilization rate is very low.[16] For example, in a national network of community health centers, only 2% of patients were screened for SDoH, and most had only one SDoH documented. The reasons for the extremely low uptake of the existing SDoH screenings are multifold, including[17–22] (**1**) existing screening tools are not automated, making them difficult to be adapted to clinical workflows, and (**2**) almost all tools were developed for universal screening but were not validated to predict specific outcomes. Furthermore, screening for individual-level SDoH items at the point of care is not only inefficient, increasing the provider documentation burden, but also inadequate given the known complex interplay of multiple SDoH.[23–26] *Figueroa et al.* called for using a PsRS approach,[27] yet the limited existing PsRS studies included predominantly individual-level SDoH and examined in small cohort studies with limited generalizability.[28–30] It is essential to consider *both* contextual (e.g., neighborhood walkability) and individual-level SDoH (e.g., if the individual has financial constraints) in one model given their known *interactions*, especially for T2D, as shown by us and others.[23,24,26,31]

The increasing availability of real-world data (RWD)[32,33]—such as electronic health records (EHRs), administrative claims, and billing data—and the rapid advancement of artificial intelligence (AI), especially machine learning (ML) techniques to analyze RWD provides an opportunity to develop novel personalized tools and generate real-world evidence for improving health outcomes. However, most studies that used ML models for clinical applications[34] did not carefully consider the inherent biases in observational RWD, such as data bias where patients of low socioeconomic status may not be well-represented in EHRs due to their limited access to healthcare.[35] A ML model naively trained on such RWD may deliver unfair outputs for racial-ethnic minority groups and socioeconomically disadvantaged individuals[35], leading to increased health disparities and inequity. Moreover, the black box nature of ML models limits their adoptions in clinical and health care applications; and explainable AI (XAI) techniques play a significant role in bridging the gap between complex ML models and human understanding.[36–38] Shapley Additive exPlanations (SHAP)[39] is increasingly used, simple tool for teasing out individual factors' contribution to the prediction, nevertheless, has limited ability to explain how the factors collectively affect the outcome with complex interactions among them, such as complex interplay among individual-level and contextual-level SDoH. Causal structure learning methods such as the classic PC algorithm[40] can learn causal relationships among the factors in the format a directed acyclic graph (DAG) from observational data, and reveal how these risk factors interact to influence outcomes, offering valuable insights into the underlying processes that drive the predictions.[41–44] Knowing what factors are important is not sufficient to make decisions, where knowing the "*treatment effect*" of a treatment or an exposure to the outcome of interest is critical to answer the "*what-if*" questions, e.g., how much hospitalization risk can we reduce, if we increase the access of healthy food—an important SDoH of T2D management. Because of the inherent bias in observational RWD, causal inference approaches such as doubly robust learning[41,44] are needed to provide accurate estimates of average treatment effect (ATE).[45]

Therefore, in this study, we aimed to develop an EHR-based ML pipeline, namely individualized polysocial risk score (iPsRS), for screening the increased social risk in hospitalization specific to T2D care and outcomes, with in-depth consideration on model fairness and explainability. We used RWD from the University of Florida Health (UF Health) EHRs and incorporated both individual-level and contextual-level SDoH for the iPsRS development. In addition, we assessed and optimized the fairness of our iPsRS model across racial-ethnic groups, sex, and age groups and identify key causal factors that can be targeted for interventions.

**Methods**

*Study design and population*

We conducted a retrospective cohort study using EHR data from the UF Health Integrated Data Repository (IDR), an enterprise data warehouse integrating different patient information systems across the UF Health system, including the Epic EHR system since 2011. UF Health provides care to more than 1 million patients with over 3 million inpatient and outpatient visits each year with hospitals in Gainesville (Alachua County), Jacksonville (Duval County), and satellite clinics in other Florida counties. In the current study, we included patients who were (1) aged 18 and older, (2) had T2D diagnosis, identified as having at least one inpatient or outpatient T2D diagnosis (using ICD-9 codes 250.x0 or 250.x2, or ICD-10 code E11) and ≥ 1 glucose-lowering drug prescription in 2015-2021 (a case finding algorithm previous validated in EHRs with a positive predictive value [PPV] >94%)[46], and (3) had least one encounter during both baseline period and the follow up year. The index date was defined as the first recorded T2D diagnosis in the UF Health IDR data. We traced back 3 years prior to the index date as the baseline period to collect predictor information and followed up for 1 year to collect outcome (i.e., hospitalization) information (**Figure 1**).

*Study outcome*

The study outcome was all-cause hospitalization within 1 year after the index date, identified using the first occurrence of an inpatient encounter during the follow-up year (**Figure 1**).

*Covariates*

Demographics and clinical characteristics

We collected patient demographic (i.e., age, sex, and race/ethnicity) and clinical information (e.g., BMI, HbA1c, Triglycerides) at the baseline period. Patients' zip codes of their residencies were collected during the baseline period for contextual-level SDoH linkage.

Individual-level SDoH extracted using natural language processing

We employed a natural language processing [47,48] pipeline that was developed by our group [49] to extract individual-level SDoH information from clinical notes in the baseline period, including education level (i.e., college or above, high school or lower, and unknown), employment (i.e., employed, unemployed, retired or disabled, and unknown), financial constraints (i.e., has financial constraints and unknown), housing stability (i.e., homeless or shelter, stable housing, and unknown), food security (i.e., having food insecurity and unknown), marital status (i.e., single, married or has partner, widow or divorce, and unknown), smoking status (i.e., ever smokers, never, and unknown), alcohol use (i.e., yes, no, and unknown), and drug abuse (i.e., yes, no and unknown).

Contextual-level SDoH through spatiotemporal linkage with the external exposome

To obtain the contextual-level SDoH, we extracted the built and social environment measures (n=114 variables) including information on food access, walkability, vacant land, neighborhood disadvantage, social capital, and crime and safety, from six well-validated sources with different spatiotemporal scales (**Supplement Table S1**) built upon our prior work.[50,51] We spatiotemporally linked these measures to each patient based on their baseline residential address (i.e., patients' 9-digit zip codes). Area-weighted averages were first calculated according to a 250 miles buffer around the centroid of each 9-digit ZIP code. Time-weighted averages were then calculated accounting for each individual's residential address.

**Overview of our ML pipeline for iPsRS**

**Figure 2** shows our overall analytics pipeline. First, we filled in any values present in the extracted data using the "unknown" for categorical variables and the mean for continuous variables. Then, we adopted balance processing techniques to manipulate the training set of the extracted data (Step 1. Preprocessing). After that, we trained a set of machine learning models by using grid search cross-validation to identify the best hyperparameters (Step 2. ML Modeling). Next, we evaluated the model prediction performance (Step 3. Performance Assessment) and utilized explainable AI and causal learning techniques to identify important SDoH contributing to hospitalizations and identify the causal relationships between the identified important SDoH and the hospitalization outcome (Step 4. Explanation). Finally, we assessed the algorithmic fairness for each model (Step

5. Fairness Assessment) and implemented a range of fairness mitigation algorithms to address the identified bias (Step 6. Fairness Mitigation).

**Data preprocessing**

We first manually categorized the variables into two types: categorical and continuous variables. We then imputed missing values using the "unknown" label for categorical variables and the mean (the average value) for continuous variables. Next, we proceeded to create dummy variables for the categorical variables and applied min-max normalization to the continuous variables.

**Machine learning model development for iPsRS**

We developed the iPsRS model for predicting hospitalizations in patients with T2D using three sets of input features: (1) individual-level SDoH only, (2) contextual-level SDoH only, and (3) individual- and contextual-level SDoH combined. Two classes of commonly used machine learning models were employed, namely linear and tree-based models. For the linear models, we adjusted a range of hyperparameters and penalty functions that can be utilized in constructing different models, including logistic regression, lasso regression, ridge regression, and ElasticNet. For the tree-based models, we selected Extreme Gradient Boosting (XGBoost), which is widely recognized as one of the best-in-class algorithms for decision-tree-based models and has shown remarkable prediction performance in a wide range of studies[52–57]. Following ML best practices, the study data set was split into a modeling set that contains T2D patients diagnosed from 2015 to 2020, and an independent testing set that covers T2D patients diagnosed in 2021. In the modeling set, we further split the samples into training, validation, and testing sets with a ratio of 7:1:2. A five-fold cross-validation grid search was executed on the training set to optimize the model parameters, and early stopping was adopted and performed on the validation set to avoid overfitting. We employed random over-sampling (ROS), random under-sampling (RUS), and under-sampling by matching on Charlson Comorbidity Index (CCI) to address data imbalance before model training. The performance of each model was evaluated by area under the receiver operating characteristic curve (AUROC), F1 score, precision, recall, and specificity.

We acquired and assigned a hospitalization risk score using the iPsRS for each patient. We then divided the ranked risk scores into 11 risk groups (i.e., deciles), enabling us to examine the one-year hospitalization rate by risk group.[58]

**Explainable AI and causal estimates**

We first utilized Shapley Additive exPlanations (SHAP)[39] – a commonly used explainable AI technique – to identify important SDoH features contributing to iPsRS predicting hospitalizations in T2D patients. Further, we used a causal discovery learning – the Mixed Graphical Models with PC-Stable (MGM-PC-Stable)[40,59–61] – to learn causal structures in directed acyclic graph (DAG) format explaining the potential causal relationships on how collectively the identified important SDoH features impact the hospitalization outcome in T2D patients.

**Algorithmic fairness assessment and optimization**

To assess the iPsRS model fairness, we adopted seven popular algorithmic fairness metrics,[35,62] including predictive parity, predictive equality (false positive rate [FPR] balance), equalized odds, conditional use accuracy equality, treatment equality, equality of opportunity (false negative rate [FNR] balance), and overall accuracy equality. We primarily focused on balancing the FNR (i.e., equality of opportunity) across racial-ethnic groups, particularly Black (NHB) and Hispanic vs. White (NHW), because hospitalization is an adverse outcome. Decreasing FNR of iPsRS means minimizing the errors (i.e., those whom the model deemed not at high risk but indeed have the risk) in early detection of social risks that can lead to hospitalization; and in terms of fairness, we want to ensure iPsRS does not have higher error rates in the disadvantaged groups (i.e., Hispanic and NHB) comparing to the reference group (i.e., NHW). As there is no universally accepted cut-off value of fairness, we considered the parity measure of 0.80-1.25 as statistically fair and highlighted values outside this range.[63]

We then employed different bias mitigation techniques to optimize the algorithmic fairness of iPsRS, including pre-process (Disparate Impact Remover[64]), in-process (Adversarial Debiasing[65]), and post-process (Calibrated Equalized Odds Postprocessing[66]) approaches, balancing the model performance between the disadvantaged groups and the reference group.

Python version 3.7 with the Python libraries Sciki-learn[67], Imbalanced-learn[68], and statsmodels[69] were used for data processing, modeling, and result analysis tasks, AI

Fairness 360[70] for model fairness mitigation tasks, and Tetrad[71] for causal structure learning.

**Results**

**Descriptive statistics of the study cohort**

Our final analysis comprised 10,192 eligible T2D patients in the cohort. **Table 1** highlights the demographics, individual-level SDoH, and key contextual-level SDoH of the study population by race-ethnicity. Overall, the mean age was 58 (± 13) years, and 58% were women. Of the cohort, 41% were enrolled in Medicare, 15% in Medicaid, 31% in private insurance, and 5.7% were uninsured. Compared with NHW patients, NHB patients were younger (54.6 vs. 58.5 years, p < 0.01) and more likely to be covered by Medicaid (41% vs. 28%, p < 0.01), while Hispanics and patients of other race-ethnicity were older (mean age of Hispanics: 61 years, other race/ethnicity: 60 years), and more likely to be women. Regarding the individual-level SDoH, it was found that 20.8% of patients were single, 58.5% were married or in a relationship, and 20.1% were widowed or divorced. Additionally, among the cohort, the percentage of smokers (40.2%) was lower than non-smokers (59.8%), the percentage of those who reported alcohol use was lower compared to those who did not use (25.8% vs 74.2%), and a similar trend was observed for drug abuse, with 4.9% of patients reporting drug abuse and 94.9% not reporting it. Concerning the contextual-level SDoH, we listed the primary contributing factors identified by both models, such as the murder rate (with an average of 0.0075), aggravated assault rate (with an average of 0.3867), and motor vehicle theft rate (with an average of 0.2348) per 100 population. According to the report by the Federal Bureau of Investigation[72], the average rates of murder and aggravated assault surpass the state average in Florida, while the motor vehicle theft rate remains below average. Among these features, crime rates within the White demographic are comparatively lower than those within other groups, including Black, Hispanics, and Others.

**iPsRS prediction model of hospitalizations in T2D patients.**

The best-performed models generated by XGBoost and ridge regression with three different sets of SDoH, including combined full SDoH, individual-level SDoH only, contextual-level SDoH only, are shown in **Figure 3**. The model included individual-level SDoH had good utility in predicting T2D hospitalization (AUC 0.70-0.71) and adding contextual-level SDoH slightly improved the model performance (AUC 0.72), while contextual-level SDoH by themselves had a suboptimal predicting utility for hospitalization (AUC 0.60-0.62).

In the testing set (i.e., 2021 data), we calculated the one-year hospitalization rates by decile of the XGBoost-generated iPsRS, adjusting for patients' demographics and clinical characteristics (i.e., age, sex, race-ethnicity, and CCI), that included both contextual and individual-level SDoH, showing an excellent utility for capturing individuals at high hospitalization risk due to SDoH (i.e., one-year hospitalization risk in the top 5% of iPsRS was 28.1%, ~13 times higher than the bottom decile, **Figure 4**). In a multiple logistic regression model, adjusting for patients' demographics and clinical characteristics, iPsRS explained 40.3% of the risk of 1-year hospitalization; and per decile increase of the iPsRS, the hospitalization risk increased by 16% (adjusted odds ratio=1.16, 95%CI 1.10-1.22).

**Explainable AI to identify important SDoH contributing to iPsRS predicting hospitalization in T2D patients**

**Figure 5-a** depicts the top 15 features ranked by SHAP values from the XGBoost. Our ridge leaner model also identified housing stability as the top predictor (**Supplement S1**). Housing stability emerged as the most predictive feature in both models, followed by employment status and insurance type. Utilizing these top 15 features, we conducted a sensitivity analysis (**Figure 5-b**), which revealed that unknown housing stability had the most effect on the outcome (i.e., lowering hospitalization risk). Conversely, patients on Medicare (i.e., insurance type: medicare) and smokers (i.e., smoking status: yes) significantly increase hospitalization risk. Additionally, we investigated the SHAP values by feature combinations (i.e., combinations of 2 or 3 features, and how the combinations contribute to the prediction). From the results presented in Supplements S2 and S3, it was evident that the joint influence of having Medicare insurance and being a smoker has a positive effect on the model's prediction of hospitalization. This is logical as older individuals who smoke are more likely to face increased risks of hospitalization. Moreover, residing in areas with elevated rates of motor theft and aggravated assault, even in a stable housing environment, is associated with heightened risks of hospitalization.

**Figure 6** displays our exploratory analysis with causal structure learning, applying MGM-PC-Stable method to build the causal DAGs of the key SDoH (i.e., 18 unique SDoH features by combining the top-15 features from both the XGBoost and ridge regression models), resulting in a causal graph with 19 nodes (i.e., 18 SDoH and the outcome) and 36 directed edges. We identified that the aggravated assault rate is closely, causally related to the hospitalization outcome, with having a direct causal connection to hospitalization. Furthermore, the rate of aggravated assault can be viewed as a common cause of both housing stability and hospitalization, forming a fork structure where housing stability and hospitalization are dependent and correlated but conditionally independent given the aggravated assault rate. This finding aligns with the insights derived from SHAP values obtained from both XGBoost and rigid leaner models, which suggests that an individual-level SDoH, housing stability, plays a significant role in T2D hospitalization, but this influence is conditioned by the contextual-level SDoH, specifically the rate of aggravated assault.

**Fairness assessment of different models.**

**Figure 9** displays the FNR curves across the racial-ethnic groups, where XGBoost (**Figure 7-a**) appears to be more fair than the linear model (**Figure 7-b**). The linear model shows a lower FNR for NHW than NHB and Hispanic groups, suggesting the model is biased towards NHB and Hispanic groups compared to NHW. The results of equality of opportunity (i.e., FNR ratios) for NHB or Hispanic vs. NHW. are shown in **Table 2**. The overall results that demonstrate all seven-fairness metrics can be found in **Supplement Tables**).

**Fairness mitigation**

To address the issue of biased predictions, we utilized three fairness mitigation algorithms and applied them to the ridge regression models. **Figure 8-a** and **Figure 8-b** illustrate the experimental results of NHB (protected group) vs. NHW (privileged group) and Hispanic vs. NHW, respectively. In these figures, the blue line represents a model that is absolutely fair, while the dotted red lines indicating the cut off thresholds we considered for a fair model between 0.8 to 1.25. Overall, DIR demonstrated better results by balancing AUROC (0.71 vs. 0.72 of the original model) and equal opportunity (FNR ratio=1.07) between the protected and privileged groups (i.e., NHB vs. NHW), presenting a promising solution to fairness mitigation compared to DIR and ADB.

**Discussion**

In this project, we develop a fair, explainable ML pipeline, namely **iPsRS,** for identifying the increased social risk of hospitalizations in patients with T2D. We used data from 10,192 patients' EHRs and incorporated both individual-level and contextual-level SDoHs. Our results demonstrated that iPSRS is a promising tool for fairly and accurately detecting patients with a higher social risk of hospitalization and providing explainable information on focal targets for future interventions.

Identifying patients' unmet social needs in health care settings is a complex task, especially when it comes to multifaceted challenges, such as the insufficient SDoH records in EHR systems, the concerns about the extra burden on providers[11,73,74] and potential harms on patients[19,21,22,75], the potential bias that exists within different subpopulations associated with SDoH, and the lack of information on effective interventions. Our automated, EHR-based iPsRS pipeline can overcome the abovementioned limitation. The further advantages have also included: 1) it considers both contextual-level and individual-level SDoH to identify the patient risk of hospitalization, taking into account the intricate interplay between them, which is comprehensive and allows for a more holistic understanding of the social risks patients face; and 2) our iPsRS assesses fairness and adjusts the model to mitigate potential biases specifically toward racial and ethnic minority groups. In fact, our additional analyses suggested that adding contextual-level SDoH may modestly improve the prediction of hospitalization in T2D compared to the individual-level SDoH-only prediction.

With multiple explainable AI and causal learning techniques, our iPsRS is able to generate interpretable outputs and has shown the ability to identify focal targets for intervention and policy programs. Specifically in our study, our SHAP value and causal discovery model consistently identified housing instability as one of the key, modifiable factors contributing to increased risk of hospitalization in T2D. In the exploratory analysis using the causal inference learning approach, we further quantified the effect of housing instability on the risk of hospitalization as well as the most vulnerable subpopulation that might be affected by housing instability. These results provide real-world evidence that we can use to design tailored SDoH-based interventions, optimizing the allocation of limited financial resources.

Another strength of our study is that we assessed the algorithmic fairness of the proposed iPsRS and mitigated the existing bias to ensure equitable prediction across racial/ethnic groups. The ridge regression model produced biased results for both NHB and Hispanic groups may cause by contextual-level SDoH. However, considering both individual-level

SDoH and contextual-level SDoH may somehow mitigate the fairness issue but the model still be skewed towards the NHW group. This important consideration was often overlooked in the existing approaches.

We consider our PsRS pipeline has important clinical implications. The current US health care system faces barriers to addressing patients' social risks essential to health. Existing SDoH screening tools and interventions have limited efficiency and effectiveness for improving T2D outcomes and health equity as most of them are not tailored to address specific conditions and outcomes (e.g., T2D), and there is insufficient evidence on effective SDoH interventions, leading to a dearth of actionable knowledge (e.g., *which SDoH should be addressed and prioritized among which individuals and their effects on T2D outcomes and disparities).* RWD and AI/ML offer the opportunity to develop innovative, digital approaches to integrate social risk management into T2D care and promote a learning health community. In this project, we addressed critical methodologic barriers, including shortcomings in existing RWD infrastructure for studying SDoH, and the need for an iPsRS approach for efficient, fair, and explainable social risk screening. Our automated algorithms can screen for unmet social needs essential to T2D care and outcomes and identify intervention targets to mitigate identified social risks. With these algorithms, we will co-design with diverse stakeholders an EHR-based individualized social risk management t platform that can integrate social risk management into diabetes care, leading to a necessary paradigm shift in US healthcare delivery.

Our study is subject to several limitations. First, the analysis conducted in our study was based on a cohort of patients with T2D in the state of Florida. This limited geographical scope may impact the generalizability of our findings to populations from other regions. However, our real-world T2D patients from Florida were highly diverse with a mixture of rural and urban populations, reflecting the demographic changes occurring across the US. Nevertheless, future research should aim to broaden the generalizability of our iPsRS through federated learning and data from different geographic regions. Second, to ensure the automated feature, we only integrated individual-level SDoH variables that were already included in the NLP extracting SDoH pipeline (SODA[49]) and thus some of the important diabetes-related factors were missing, such as stress. We will continue developing NLP pipelines for expanding the list of SDoH extraction and updating our iPsRS model. Third, we based on ML practices to select and tune the proposed iPsRS, hence the searching space of models and hyperparameters is constrained. We plan to utilize AutoML pipelines to enhance model accuracy and reliability, while simultaneously minimizing the time and resources required to develop the next-generation model.

In this project, we developed an ML-based method, namely iPsRS, for identifying the increased social risk of hospitalizations in patients with T2D. Our iPsRS has been shown as a promising tool to identify patients' unmet social needs that are associated with an increased risk of adverse health outcomes. The iPsRS have the great potential to be integrated into EHR systems and clinical workflow and eventually augment current screening programs for SDoH to provide physicians with an efficient and effective tool to address SDoH in clinical settings.

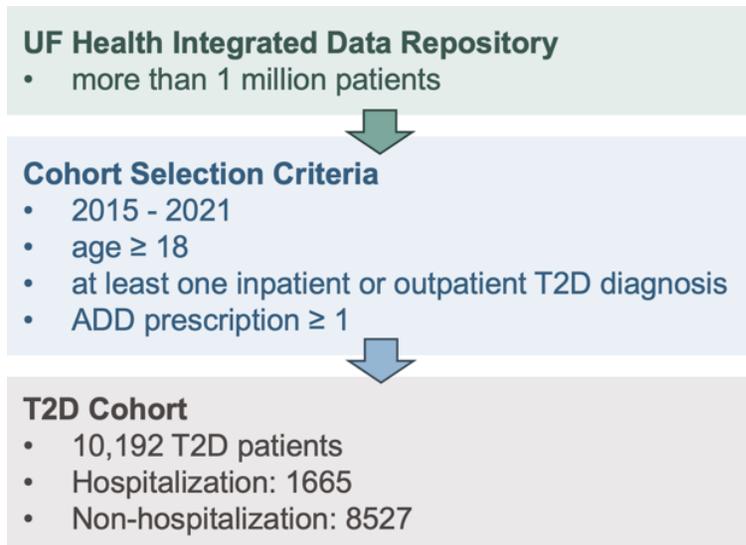
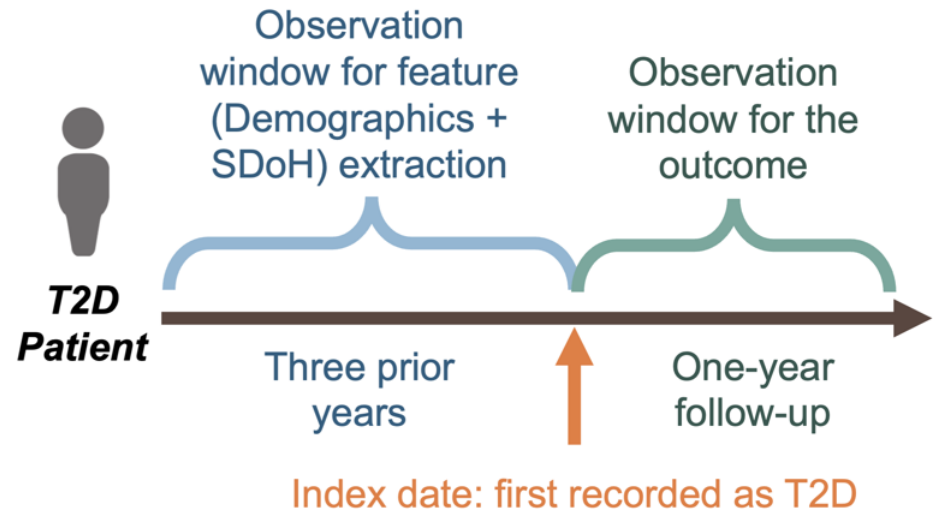

**Figure 1** Processing workflow of the University of Florida integrated data repository type 2 diabetes cohort and the patient timeline.

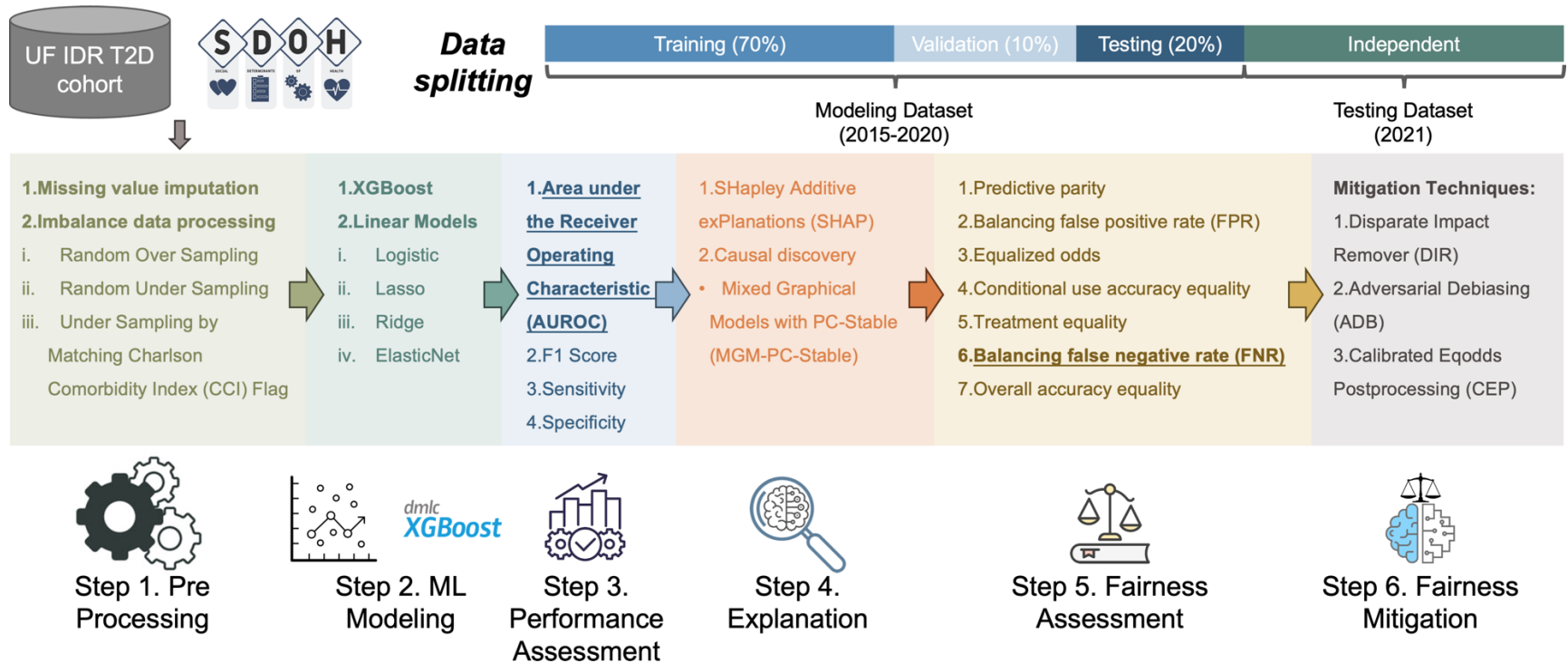

**Figure 2** Data analytics pipeline.

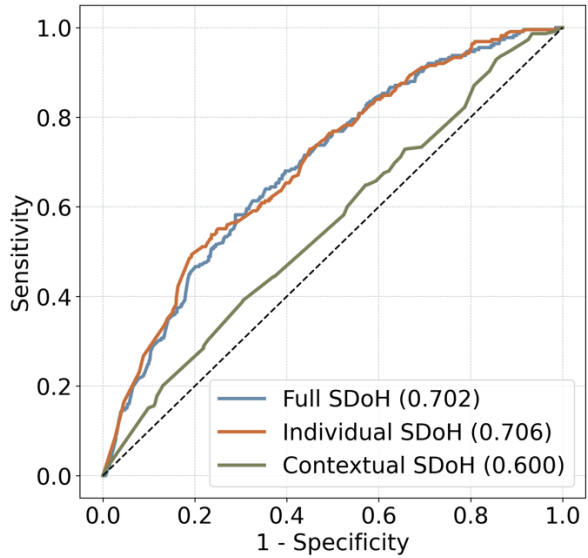 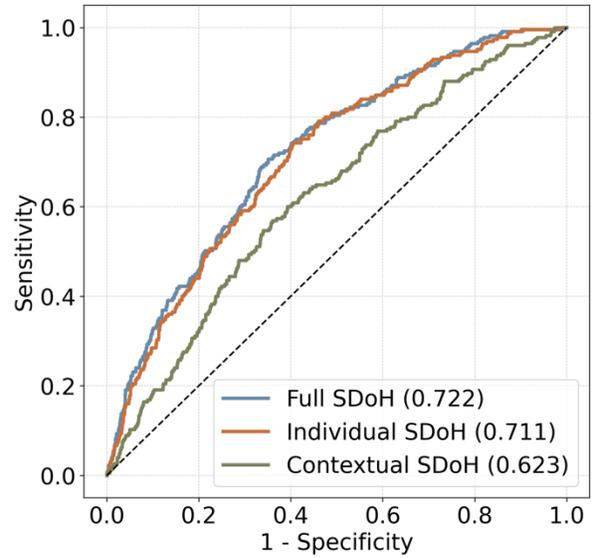

(a) XGBoost  (b) Ridge regression

**Figure 3** Model performance assessment of XGBoost and ridge regression.

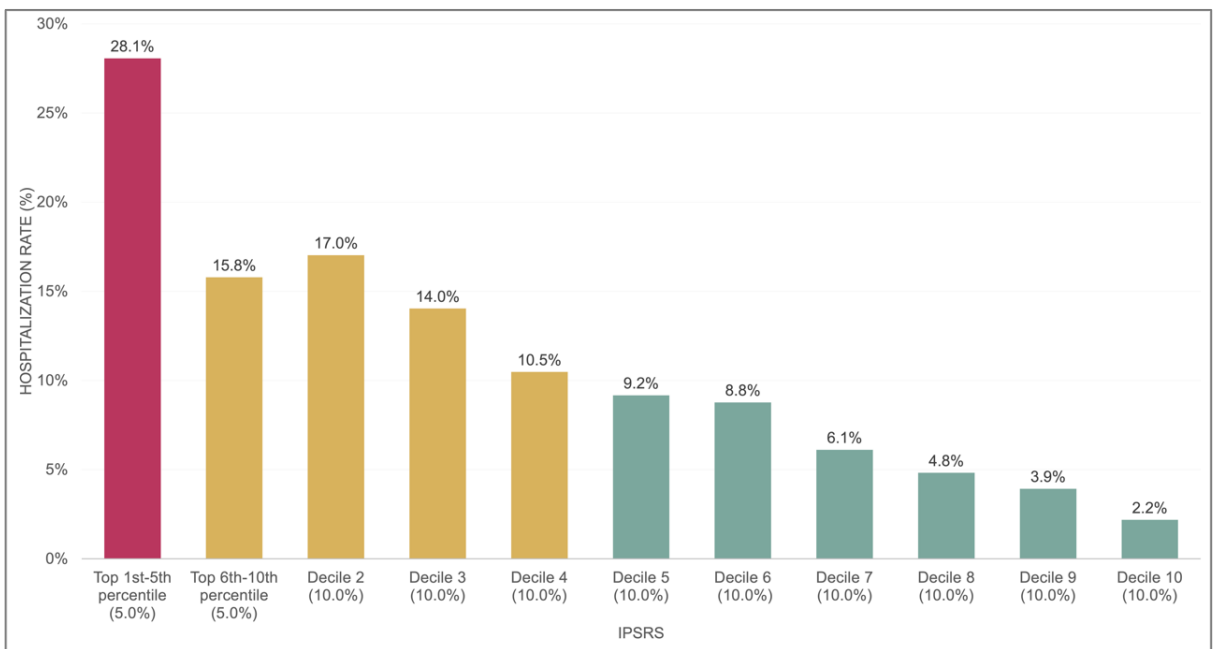

**Figure 4** One-year hospitalization risk by iPsRS decile.

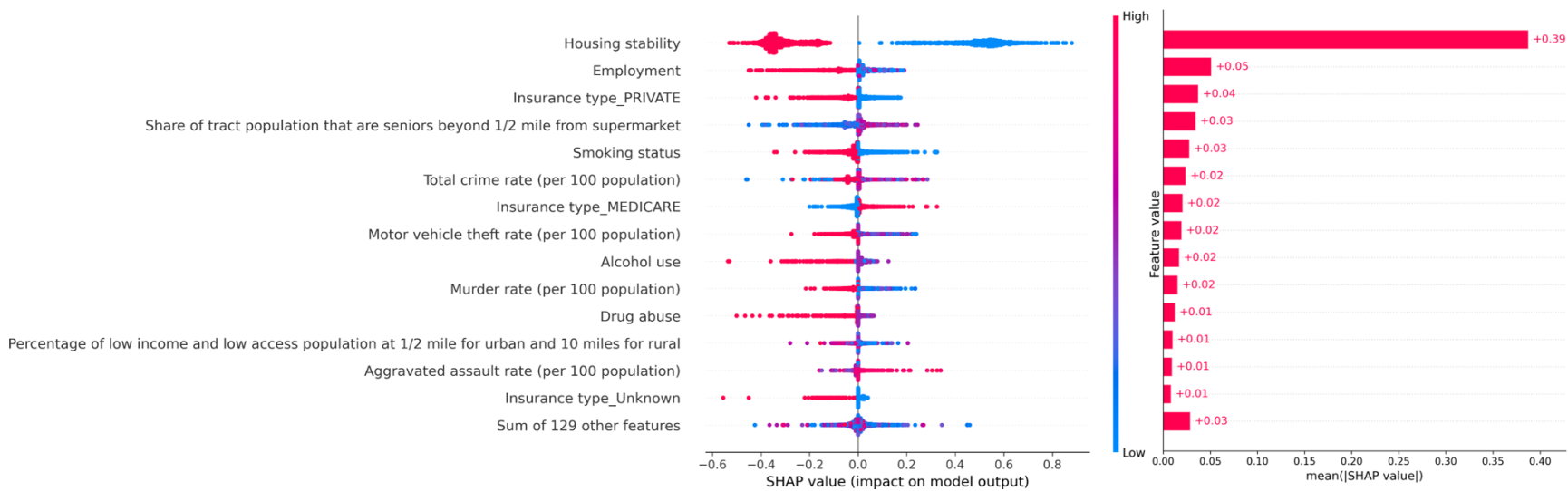

(a) SHAP values from the original XGBoost.

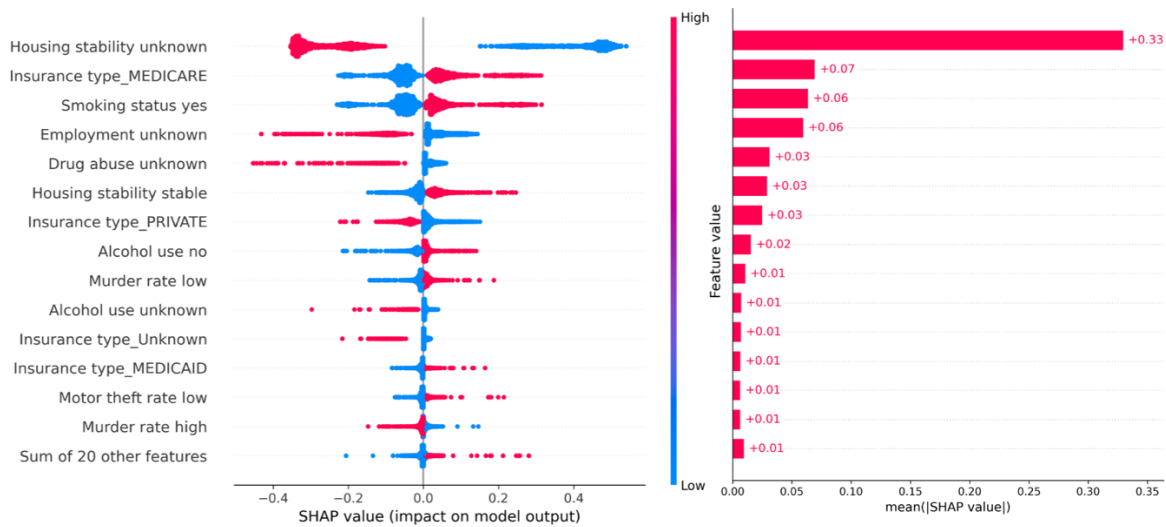

(b) SHAP values from the sensitivity analysis (Using the top 15 features from the original XGBoost to build a prediction model). We dichotomized the continuous contextual-SDoH, such as the murder rate and motor theft rate, based on the online crime data explorer provided by the Federal Bureau of Investigation[1].

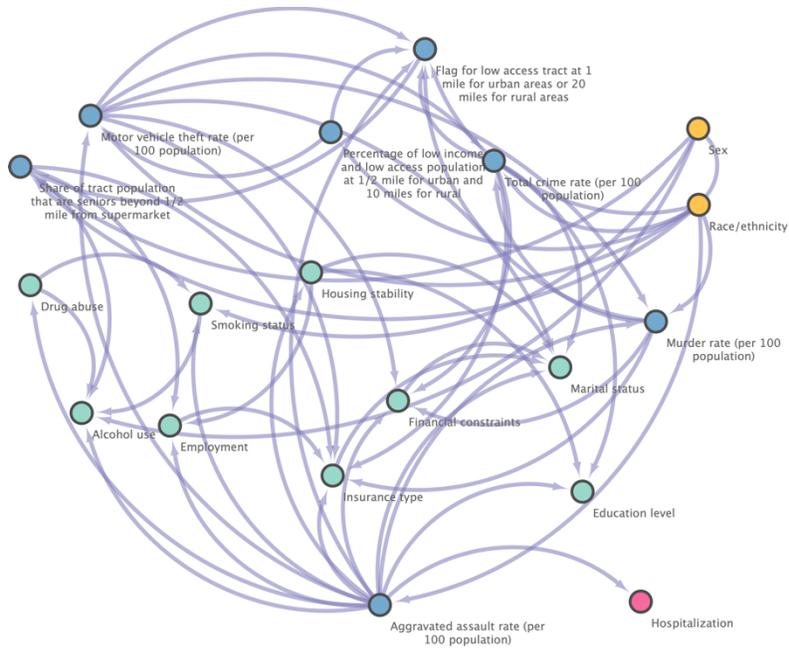

**Figure 6.** Exploratory analysis with causal discovery learning. Causal graph generated by MGM-PC-Stable in the independent testing set. The yellow nodes present demographics, blue nodes stand for contextual-level SDoH and green nodes mean the individual-level SDoH, and the pink node indicates the outcome.

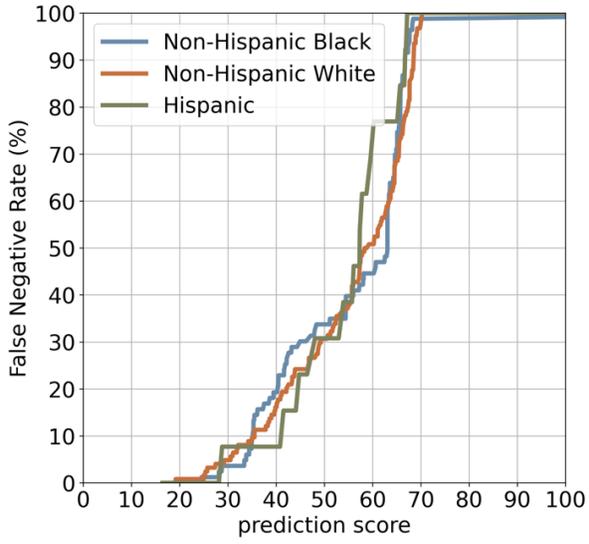 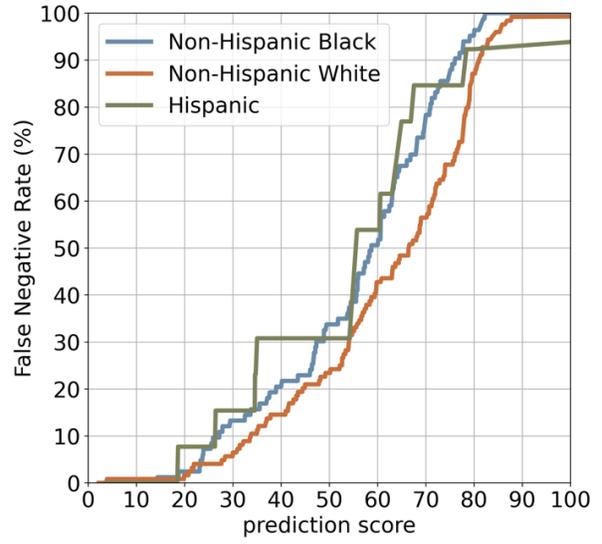

(a) XGBoost  (b) Ridge regression

**Figure 7** False negative rate (FNR) curve between different populations.

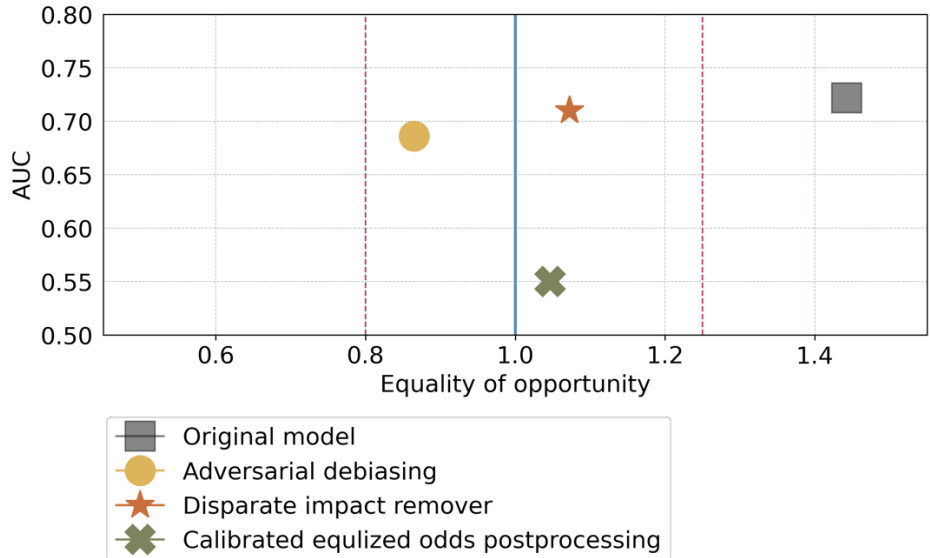

**(a)** Mitigation results on the NHB vs NHW. CEP had the best fairness mitigation ability but led to a drastic decrease in model performance from 0.7220 to 0.5501, measured by AUROC, which is unacceptable. DIR and ADB resulted in an acceptable decrease in prediction performance, particularly with DIR's AUROC decreasing from 0.7220 to 0.7100.

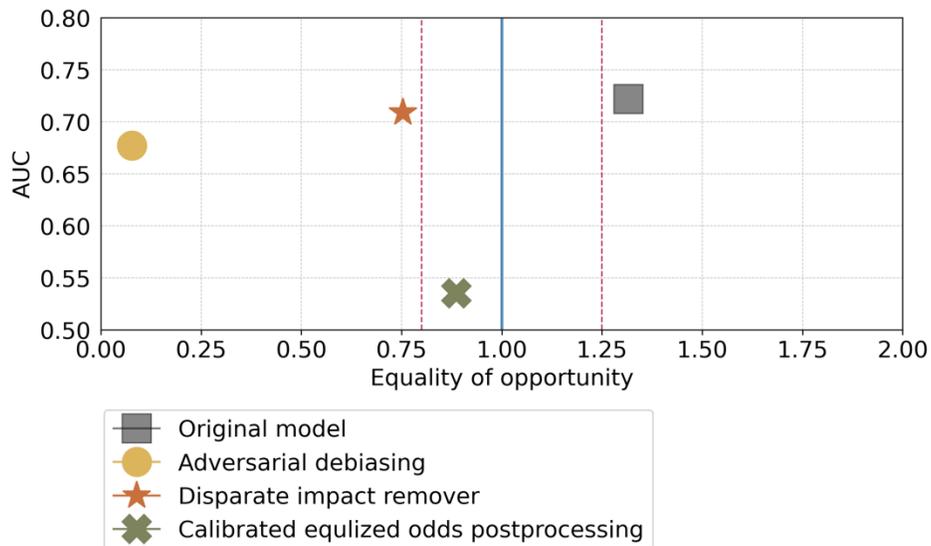

**(b)** Mitigation results on the Hispanic vs NHW. DIR and ADB struggled to handle the fairness mitigation. These methods turned to favoritism towards the protected group (Hispanic), resulting in biased predictions for the NHW group.

**Figure 8.** NHB (protected group) vs. NHW (privileged group) and Hispanic vs. NHW, respectively. The ideally fair line is represented by the blue line, while the range of statistically fair is shown by the red dots. the ridge regression model initially fell outside the range of statistically fair but became fairer when we employed the fairness mitigation methods CEP, DIR, and ADB, resulting in equal opportunity regarding FNR raito.

23  **Table 1** Summary of demographic, individual-level SDoH and key contextual-level SDoH of the study
24  population.

|  | Overall (n=10192) | NHW (n=5133) | NHB (n=4011) | Hispanics (n=495) | Others (n=553) | p-value |
|---|---|---|---|---|---|---|
| **Age** | 58.45 | 60.19 | 56.39 | 55.95 | 59.42 | 0.0049 |
| **Sex** |  |  |  |  |  | 0.0018 |
| *Male* | 4267 (41.9%) | 2470 (48.1%) | 1330 (33.2%) | 212 (42.8%) | 255 (46.1%) |  |
| *Female* | 5925(58.1%) | 2663 (51.9%) | 2681 (66.8%) | 283 (57.2%) | 298 (53.9%) |  |
| **Race/ethnicity** |  |  |  |  |  | <0.001 |
| *NHB* | 4011 (39.4%) | 0 (0.0%) | 4011 (100.0%) | 0 (0.0%) | 0 (0.0%) |  |
| *NHW* | 5133 (50.4%) | 5133 (100.0%) | 0 (0.0%) | 0 (0.0%) | 0 (0.0%) |  |
| *Hispanics* | 495 (4.9%) | 0 (0.0%) | 0 (0.0%) | 495 (100.0%) | 0 (0.0%) |  |
| *Others* | 553 (5.4%) | 0 (0.0%) | 0 (0.0%) | 0 (0.0%) | 553 (100.0%) |  |
| **Insurance type** |  |  |  |  |  | <0.001 |
| *Medicare* | 4183 (41.0%) | 2214 (43.1%) | 1610 (40.1%) | 170 (34.3%) | 189 (34.2%) |  |
| *Private* | 3169 (31.1%) | 1663 (32.4%) | 1144 (28.5%) | 148 (29.9%) | 214 (38.7%) |  |
| *Medicaid* | 1511 (14.8%) | 558 (10.9%) | 804 (20.0%) | 97 (19.6%) | 52 (9.4%) |  |
| *Nopay* | 579 (5.7%) | 228 (4.4%) | 285 (7.1%) | 38 (7.7%) | 28 (5.1%) |  |
| *Unknown* | 537 (5.3%) | 362 (7.1%) | 84 (2.1%) | 32 (6.5%) | 59 (10.7%) |  |
| *Others* | 213 (2.1%) | 108 (2.1%) | 84 (2.1%) | 10 (2.0%) | 11 (2.0%) |  |
| **Marites status** |  |  |  |  |  | <0.001 |
| *Single* | 2116 (20.8%) | 743 (14.5%) | 1221 (30.4%) | 80 (16.2%) | 72 (13.0%) |  |
| *Married or has partner* | 3570(35.0%) | 2073 (40.4%) | 1069 (26.7%) | 179(36.2%) | 249 (45.0%) |  |
| *Widow or divorce* | 2050 (20.1%) | 888 (17.3%) | 1052 (26.2%) | 65 (13.1%) | 45 (8.1%) |  |
| *Unknown* | 2456 (24.1%) | 1429 (27.8%) | 669 (16.7%) | 171 (34.5%) | 187 (33.8%) |  |
| **Smoking status** |  |  |  |  |  | <0.001 |
| *Ever smokers* | 4096 (40.2%) | 2331 (45.4%) | 1473 (36.7%) | 149 (30.1%) | 143 (25.9%) |  |
| *Never* | 5588 (54.8%) | 2525 (49.2%) | 2380 (59.3%) | 321 (64.8%) | 362 (65.5%) |  |
| *Unknown* | 508(5.0%) | 277(5.4%) | 158 (3.9%) | 25(5.1%) | 48(8.7%) |  |
| **Alcohol use** |  |  |  |  |  | <0.001 |
| *Yes* | 2631 (25.8%) | 1381 (26.9%) | 1012 (25.2%) | 123 (24.8%) | 115 (20.8%) |  |
| *No* | 6650(65.2%) | 3223(62.8%) | 2737(68.2%) | 325 (65.7%) | 365 (66.0%) |  |
| *Unknown* | 911 (9.0%) | 529 (10.3%) | 262 (6.5%) | 47(9.5%) | 73(13.2%) |  |
| **Drug abuse** |  |  |  |  |  | <0.001 |
| *Yes* | 500 (4.9%) | 225 (4.4%) | 253 (6.3%) | 16 (3.2%) | 6 (1.1%) |  |
| *No* | 8487 (83.3%) | 4218 (82.2%) | 3409 (85.0%) | 417 (84.2%) | 443 (80.1%) |  |
| *Unknown* | 1205 (11.8%) | 690 (13.4%) | 349 (8.7%) | 62(12.5%) | 104 (18.8%) |  |
| **Education level** |  |  |  |  |  | <0.001 |
| *College or above* | **978 (9.6%)** | **518 (10.1%)** | **376 (9.4%)** | 38 (7.7%) | 46 (8.3%) |  |
| *High school or lower* | **1110 (10.9%)** | **461 (9.0%)** | **563 (14.0%)** | 50 (10.1%) | 36 (6.5%) |  |
| *Unknown* | **8104 (79.5%)** | **4154 (80.9%)** | **3072 (76.6%)** | 407 (82.2%) | 471 (85.2%) |  |
| **Employment** |  |  |  |  |  | <0.001 |
| Employed | 3996 (39.2%) | 2078 (40.5%) | 1489 (37.1%) | 207 (41.8%) | 222 (40.1%) | |
| Unemployed | 1439 (14.1%) | 570 (11.1%) | 760 (18.9%) | 57 (11.5%) | 52 (9.4%) | |
| Retired or disabled | 1948 (19.1%) | 1017 (19.8%) | 782 (19.5%) | 68 (13.7%) | 81 (14.6%) | |
| Unknown | 2809 (27.6%) | 1468 (28.6%) | 980 (24.4%) | 163 (32.9%) | 198 (35.8%) | |
| **Housing stability** | | | | | | <0.001 |
| Homeless or shelter | 80 (0.8%) | 32 (0.6%) | 44 (1.1%) | 3 (0.6%) | 1 (0.2%) | |
| Stable housing | 4215 (41.4%) | 1971 (38.4%) | 1933 (48.2%) | 160 (32.3%) | 151 (27.3%) | |
| Unknown | 5897 (57.9%) | 3130 (61%) | 2034 (50.7%) | 332 (67.1%) | 401 (72.5%) | |
| **Food security** | | | | | | <0.001 |
| Having food insecurity | 7052 (69.2%) | 3416 (66.5%) | 2982 (74.3%) | 300 (60.6%) | 354 (64.0%) | |
| Unknown | 3140 (30.8%) | 1717 (33.5%) | 1029 (25.7%) | 195 (39.4%) | 199 (36.0%) | |
| **Financial constraints** | | | | | | 0.0092 |
| Has financial constraints | 5172 (50.7%) | 2386 (46.5%) | 2323 (57.9%) | 216 (43.6%) | 247 (44.7%) | |
| Unknown | 5020 (49.3%) | 2747 (53.5%) | 1688 (42.1%) | 279 (56.4%) | 306 (55.3%) | |
| **Percentage of low income and low access population at 1/2 mile for urban and 10 miles for rural** | 0.2625 (0.1965) | 0.1944 (0.1733) | 0.3528 (0.1946) | 0.2579 (0.1740) | 0.2442 (0.1685) | 0.1708 |
| **Share of tract population that are seniors beyond 1/2 mile from supermarket** | -0.1661 (0.0949) | -0.1635 (0.1035) | -0.1669 (0.0831) | -0.1734 (0.0837) | -0.1779 (0.1000) | < 0.001 |
| **Murder rate (per 100 population)** | 0.0075 (0.0043) | 0.0064 (0.0040) | 0.0089 (0.0041) | 0.0076 (0.0041) | 0.0074 (0.0044) | < 0.001 |
| **Aggravated assault rate (per 100 population)** | 0.3867 (0.1365) | 0.3767 (0.1704) | 0.3980 (0.0753) | 0.3994 (0.1489) | 0.3858 (0.1060) | < 0.001 |
| **Motor vehicle theft rate (per 100 population)** | 0.2348 (0.0882) | 0.2042 (0.0921) | 0.2718 (0.0684) | 0.2420 (0.0785) | 0.2440 (0.0794) | < 0.001 |
| **Flag for low access tract at 1 mile for urban areas or 20 miles for rural areas counts** | | | | | | < 0.001 |
| Yes | 4630 (45.4%) | 2091 (40.7%) | 2031 (50.6%) | 253 (51.1%) | 306 (55.3%) | |
| No | 5562 (54.6%) | 3042 (59.3%) | 1980 (49.4%) | 242 (48.9%) | 247 (44.7%) | |

Table 2 Statistical parity (equal opportunity) by different models on various feature sets.

| Black & White | Full SDoH | Individual SDoH | Contextual SDoH |
|---|---|---|---|
| Xgboost | 1.03 | 0.98 | 1.24 |
| Ridge regression | 1.44 | 1.18 | 1.45 |
| Hispanic & White | Full SDoH | Individual SDoH | Contextual SDoH |
| Xgboost | 1.22 | 1.00 | 1.63 |
| Ridge regression | 1.32 | 1.73 | 2.12 |